\documentclass[accepted, startpage]{uai2023} 

\usepackage[american]{babel}

\usepackage{natbib} 
\usepackage{mathtools} 
\usepackage{booktabs} 

\usepackage{graphicx}
\usepackage{bm}
\usepackage{algorithm}
\usepackage[noend]{algpseudocode}
\usepackage{multirow}
\usepackage{xcolor}
\usepackage[normalem]{ulem}
\usepackage[accsupp]{axessibility}
\usepackage{pifont}
\newcommand{\cmark}{\ding{51}}
\newcommand{\xmark}{\ding{55}}
\newcommand{\textun}[1]{\underline{#1}}
\newcommand{\eg}{e.g.}
\newcommand{\etc}{etc.}
\newcommand{\ie}{i.e.}
\newcommand{\wrt}{w.r.t.}

\def\eqref#1{equation~\ref{#1}}

\def\1{\bm{1}}
\newcommand{\train}{\mathcal{D_{\mathrm{train}}}}

\newcommand{\test}{\mathcal{D_{\mathrm{test}}}}

\def\vzero{{\bm{0}}}

\def\vmu{{\bm{\mu}}}
\def\vbeta{{\bm{\beta}}}
\def\vlambda{{\bm{\lambda}}}
\def\vtheta{{\bm{\theta}}}

\def\va{{\bm{a}}}

\def\ve{{\bm{e}}}

\def\vu{{\bm{u}}}

\def\vx{{\bm{x}}}
\def\vy{{\bm{y}}}
\def\vz{{\bm{z}}}

\def\mI{{\bm{I}}}
\def\mJ{{\bm{J}}}

\def\mU{{\bm{U}}}

\def\mSigma{{\bm{\Sigma}}}

\DeclareMathAlphabet{\mathsfit}{\encodingdefault}{\sfdefault}{m}{sl}
\SetMathAlphabet{\mathsfit}{bold}{\encodingdefault}{\sfdefault}{bx}{n}

\def\sN{{\mathbb{N}}}

\newcommand{\pdata}{p}
\newcommand{\ptrain}{\hat{p}}

\newcommand{\E}{\mathbb{E}}

\DeclareMathOperator*{\argmax}{arg\,max}

\newif\ifreport
\reporttrue

\title{Concurrent Misclassification and Out-of-Distribution Detection for \\ Semantic Segmentation via Energy-Based Normalizing Flow}

\author[1]{\href{mailto:<denis.gudovskiy@us.panasonic.com>?Subject=Your UAI 2023 paper}{Denis Gudovskiy}{}}
\author[2]{Tomoyuki Okuno}
\author[2]{Yohei Nakata}
\affil[1]{%
	Panasonic AI Lab, Mountain View, CA, USA
}
\affil[2]{%
	Panasonic Holdings Corporation, Osaka, Japan
}

\begin{document}
\maketitle

\begin{abstract}
Recent semantic segmentation models accurately classify test-time examples that are similar to a training dataset distribution. However, their discriminative closed-set approach is not robust in practical data setups with distributional shifts and out-of-distribution (OOD) classes. As a result, the predicted probabilities can be very imprecise when used as confidence scores at test time. To address this, we propose a generative model for concurrent in-distribution misclassification (IDM) and OOD detection that relies on a normalizing flow framework. The proposed flow-based detector with an energy-based inputs (FlowEneDet) can extend previously deployed segmentation models without their time-consuming retraining. Our FlowEneDet results in a low-complexity architecture with marginal increase in the memory footprint. FlowEneDet achieves promising results on Cityscapes, Cityscapes-C, FishyScapes and SegmentMeIfYouCan benchmarks in IDM/OOD detection when applied to pretrained DeepLabV3+ and SegFormer semantic segmentation models.
\end{abstract}

\section{Introduction}\label{sec:intro}
Test-time robustness is one of the most wanted yet missing properties in current machine learning (ML) models when they are applied to decision-critical computer vision applications~\citep{unsolved-ml}. Typically, ML-based models achieve high average accuracy metrics only for test-time data that are similar to a labeled training dataset distribution with a predefined set of categories. However, a test-train distributional shift and a novel open-set categories can significantly decrease accuracy~\citep{croce2021robustbench}. 

\begin{figure}[t]
	\centering
	\includegraphics[width=0.92\columnwidth]{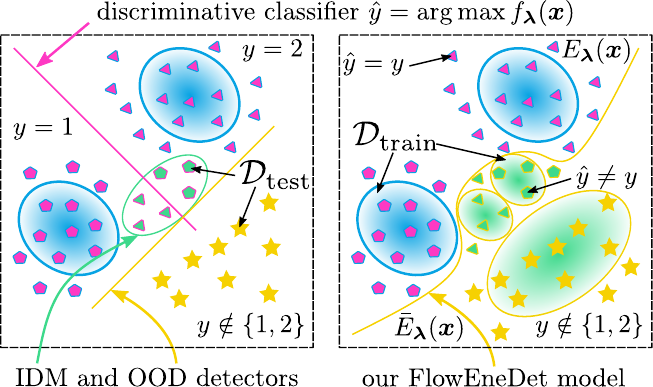}
	\caption{A discriminative model $f_{\vlambda}(\vx)$ is trained to predict segmentation classes $\hat{y}$ for images $\vx$ using an empirical dataset $\train$ (blue ovals) with a closed-set labels $y \in \{1,2\}$. However, an open-world data $\test$ (stars, triangles \etc) can contain out-of-distribution (OOD) classes $\left( y \notin \{1,2\} \right)$ and in-distribution misclassified (IDM) predictions $\left( \hat{y} \ne y \right)$. Conventional approaches (left) aim either IDM or OOD detection. Our FlowEneDet (right) is a generative normalizing flow model that estimates likelihoods of correctly classified in-distribution data (purple positives) as well as IDM (green negatives) and OOD (yellow negatives) samples. We achieve this by modeling distributions of a scalar free energy score $E_{\vlambda}(\vx)$ for positives and an opposite $\bar{E}_{\vlambda}(\vx)$ for negatives using $\train$ (green ovals).}
	\label{fig:problem}
\end{figure}


We sketch this scenario with a toy example in Figure~\ref{fig:problem}. Here, a discriminative task model $f_{\vlambda}(\vx)$ misclassifies test examples (green triangles and pentagons), and assigns wrong closed-set class predictions to novel categories (yellow stars) due to lack of coverage in a training dataset (blue ovals). In-distribution misclassification (IDM) and out-of-distribution (OOD) detection are test-time approaches for the above problem. Conventional IDM and OOD detectors estimate confidence scores for the classifier predictions as shown in Figure~\ref{fig:problem} (left). OOD detector separates a distribution of unknown categories $(y \notin \{1,2\})$ from a distribution of known categories $(y \in \{1,2\})$ using a threshold~\citep{morteza2022provable}. IDM detection aims to identify correctly ($\hat{y} = y$ positives) and incorrectly ($\hat{y} \ne y$ negatives) classified in-distribution data~\citep{ramalho2020density}.

Existing detectors experiment with either IDM or, more often, OOD detection. Our analysis shows that IDM and OOD detection objectives have common root causes and, hence, can be addressed concurrently. We approach both objectives by explicitly modeling distributions of free energy function for positives and IDM/OOD negatives as shown in Figure~\ref{fig:problem} (right). Inspired by~\citet{djurisic2023extremely}, we explicitly learn \textit{what a trained discriminative model knows and what it doesn't know} from the empirical training dataset. To accomplish this, we propose a low-complexity generative normalizing flow model (FlowEneDet) for concurrent IDM/OOD detection, which is trained on top of a fixed (task-pretrained) discriminative semantic segmentation model. In summary, our contributions are as follows:
\begin{itemize}
	\item We derive a low-complexity flow-based model to estimate exact likelihoods of free energy both for positives and negatives from the training dataset.
	\item We tailor it for semantic segmentation application as a compact and stable 2D Glow-like~\citep{NEURIPS2018_d139db6a} architecture that employs both the logit- and latent-space spatial context information.
	\item FlowEneDet achieves promising results on IDM/OOD benchmarks~\citep{michaelis2020benchmarking, fishyscapes, segmentmeifyoucan} for the  task-pretrained setup\footnote{\href{https://github.com/gudovskiy/flowenedet}{Our code is available at github.com/gudovskiy/flowenedet}}.
\end{itemize}

\begin{table*}[ht]
	\caption{A landscape of IDM/OOD detectors for semantic segmentation. Symbols indicate: \cmark~for "yes", \xmark~for "no", and $\dagger$ for a possible extension. We categorize methods by: discriminative or generative type, intact task mIoU accuracy (no retraining setup), IDM detection, extra network for detection, inference speed (time for detection is lower than the segmentation), source of negatives such as in-domain data (void class, misclassified pixels) or proxy dataset to emulate OOD distribution.}
	\label{tab:landscape}
	\centering
	\small
	\begin{tabular}{c|c|c|c|c|c|c|c}
		\toprule
		\multirow{2}{*}{Method} & \multirow{2}{*}{Type} & \multirow{2}{*}{\shortstack{Intact mIoU, \\ no retraining}} & \multirow{2}{*}{\shortstack{IDM \\ detection}} & \multirow{2}{*}{\shortstack{Extra det. \\ network}} & \multirow{2}{*}{\shortstack{Fast \\ inference}} & \multirow{2}{*}{\shortstack{In-domain \\ negative data}} & \multirow{2}{*}{\shortstack{Extra OOD \\ negative data}} \\
		& & & & & & \\
		\midrule
		MSP, MLG, ENE, SML               & disc. & \cmark & \cmark             & \xmark & \cmark & \xmark & \xmark \\
		ODIN, MCD                        & disc. & \cmark & \cmark             & \xmark & \xmark & \xmark & \xmark \\
		Mahalanobis distance             & disc. & \cmark & \cmark             & \xmark & \cmark & \xmark & \xmark \\
		SynthCP, Image Resynthesis       & gen.  & \cmark & \cmark             & \cmark & \xmark & \xmark & \xmark \\
		SynBoost                         & gen.  & \cmark & \xmark$^{\dagger}$ & \cmark & \xmark & \cmark & \xmark \\
		ObsNet                           & disc. & \cmark & \cmark             & \cmark & \xmark & \cmark & \xmark \\
		Flow emb. density                & gen.  & \cmark & \xmark$^{\dagger}$ & \cmark & \xmark & \xmark & \xmark \\
		\textbf{FlowEneDet (ours)}       & gen.  & \cmark & \cmark & \cmark    & \cmark & \cmark & \xmark \\
		\midrule
		NFlowJS                             & disc. & \xmark & \xmark             & \xmark & \cmark & \cmark~(sampled) & \xmark \\
		Meta-OOD                            & disc. & \xmark & \xmark             & \xmark & \cmark & \cmark & \cmark~(COCO) \\
		PEBAL                               & disc. & \xmark & \xmark             & \xmark & \cmark & \xmark & \cmark~(COCO) \\
		DenseHybrid                         & disc. & \xmark & \xmark             & \cmark & \cmark & \xmark & \cmark~(ADE20K) \\
		GMMSeg                              & gen.  & \xmark & \xmark             & \xmark & \cmark & \xmark & \xmark \\
		\bottomrule
	\end{tabular}
\end{table*}

\section{Related Work}\label{sec:related}
IDM and OOD detection is an active area of research for many ML-centric applications. We survey and compare a line of research that estimates categorical classifier's confidence scores for semantic segmentation in Table~\ref{tab:landscape}.

Several popular methods estimate confidence scores at the output of a task classifier. These include a maximum of softmax probabilities (MSP)~\citep{hendrycks17baseline} or unnormalized logits (MLG)~\citep{hendrycks2020scaling}, standardized logits (SML)~\citep{Jung_2021_ICCV}, an energy-based detection (ENE)~\citep{NEURIPS2020_f5496252}, and ODIN~\citep{liang2018enhancing}. The latter has higher complexity due to test-time gradient perturbations. In-distribution scores in such methods are often accurate in the proximity of train data distribution due to the task's Kullback-Leibler (KL) divergence objective, but less accurate for OOD data~\citep{NEURIPS2019_8ca01ea9}.

\citet{ken} propose an uncertainty-based detector that relies on approximate Bayesian inference (MCD). A notion of uncertainty can be viewed as an alternative way to define low confidence. MCD is implemented using forward passes at test-time for a task model with dropout layers and a scoring function. Unfortunately, its complexity scales linearly with the number of passes without approximation methods~\citep{Postels_2019_ICCV}, and the dropout layer's configuration is sensitive to heuristic hyperparameters.

\citet{NEURIPS2018_abdeb6f5} model data distributions using Gaussian discriminant analysis in the task's latent-space, and employ Mahalanobis distance as a confidence score. The above SML improves OOD accuracy using a similar approach, but operates in low-dimensional logit-space. Their main drawback is the assumption of Gaussian prior, which can be inaccurate in multi-label classification~\citep{kamoi2020mahalanobis}.

Variational autoencoders~\citep{baur2018deep} and generative adversarial networks (GANs) can be used to implement reconstruction-based detectors by training a dedicated generative model at the expense of higher complexity (Image Resynthesis by~\citet{Lis_2019_ICCV} and SynthCP by~\citet{xia2020synthesize}). Then, a test-time difference between an input image and a generated image is a proxy of the confidence score. Unlike normalizing flows~\citep{rezende15}, such models cannot estimate the exact data likelihoods and can be unreliable due to the tendency of capturing semantically-irrelevant low-level correlations~\citep{nalisnick2018do}. SynBoost~\citep{Di_Biase_2021_CVPR} addresses the latter by combining GAN sampling with other non-parametric methods.

\citet{besnier2021triggering} propose a dedicated observer (ObsNet) that exactly mirrors the task model architecture. It is trained to predict misclassifications using binary cross-entropy loss and adversarial attacks. Therefore, ObsNet is an improved discriminative model similar to a simple OOD detection head in~\citep{bevandic}. Unlike it, our FlowEneDet is a theoretically more robust generative model that processes low-complexity scalar free energy scores.

\citet{fishyscapes} introduce a relatively high complexity latent-space flow-based density estimator (flow emb. density) trained using marginal likelihood objective with the pretrained task model. Unlike it, FlowEneDet has significantly lower complexity, and, importantly, a more advanced distributional model that supports joint likelihood estimation for positives and negatives. Though not implemented, this density estimator and SynBoost~\citep{Di_Biase_2021_CVPR} without task retraining can be used for IDM detection.

Lastly, we contrast the above IDM/OOD detectors from OOD-only methods at the bottom of Table~\ref{tab:landscape}. The latter retrain all task model parameters (NFlowJS~\citep{nflowjs}, Meta-OOD~\citep{Chan_2021_ICCV}, DenseHybrid~\citep{grcic22eccv}, GMMSeg~\citep{liang2022gmmseg}) or its subset (PEBAL~\citep{pebal}). GMMSeg does not rely on an outlier exposure~\citep{wang2023outofdistribution}, while NFlowJS is trained with the sampled negatives. Others emulate OOD distribution by a proxy data such as COCO~\citep{lin2014microsoft} or ADE20K~\citep{Zhou_2017_CVPR} with augmentations~\citep{li2021cutpaste}. Though such methods currently achieve state-of-the-art results in OOD-only detection, they bear several major limitations such as: lack of IDM detection, inability to extend already deployed task models, and a certain degradation in tasks' in-domain segmentation accuracy. We compare FlowEneDet to these baselines on OOD-only benchmarks.

\section{Theoretical Background}\label{sec:theory}
\subsection{Limitations of Conventional Closed-Set Discriminative Models}\label{subsec:discussion}
Let $(\vx, y)$ be an input-label pair where a vector $\vx$ is an input image and a closed-set scalar label $y \in \{1, \ldots, C\}$ has $C$ classes. Then, a conventional discriminative model $f_{\vlambda}(\vx)$ from Figure~\ref{fig:problem} is optimized using a supervised training dataset $\train=\{(\vx_i,y_i)\}_{i\in\sN}$ of size $N$ with an empirical risk minimization objective expressed by
\begin{equation} \label{eq:p1}
	\mathcal{L}(\vlambda) = \frac{1}{N} \sum_{i \in \sN} L(y_i, \mathrm{softmax} f_{\vlambda}(\vx_i)),
\end{equation}
where $L(\cdot)$ is a loss function, $\vlambda$ is the vector of parameters. The classifier's test-time prediction $\hat{y} = \argmax \hat{\vy}$, where the vector of unnormalized logits $\hat{\vy} = f_{\vlambda}(\vx) \in \mathbb{R}^{C}$.

Typically, discriminative models minimize KL divergence $D_{KL} \left[ \pdata (\vx,y) \| p_{\vlambda} (\vx,y) \right]$ between, correspondingly, the joint data and model probability density functions in the~(\ref{eq:p1}) loss. However, the underlying $\pdata (\vx,y)$ is a-priori unknown for test data and it is \textit{approximated by the empirical training set} $\train$ with $\ptrain (\vx,y) = \ptrain (y | \vx) \ptrain (\vx) $ density function. As shown in~\citep{Gudovskiy_2020_CVPR}, the KL divergence for (\ref{eq:p1}) with one-hot labels $y$ \ie~the cross-entropy loss can be equivalently derived with these notations as
\begin{equation} \label{eq:p2}
\E_{\vx \sim \ptrain (\vx)} D_{KL} \left[ \ptrain (y | \vx) \| p_{\vlambda} (y | \vx) \right] = - \frac{1}{N} \sum_{i \in \sN} \log p_{\vlambda} (y_i | \vx_i).
\end{equation}

Hence, the discriminative approach is limited to modeling conditional density $p_{\vlambda} (y | \vx)$, where inputs are sampled as $\vx \sim \ptrain (\vx)$ and labels $y$ are from the closed set.

\subsection{Motivation and Problem Statement for Concurrent IDM/OOD Detection}\label{subsec:motivation}
Conventional OOD detection data setup assumes an in-distribution $\pdata (\vx)$ and an out-of-distribution $\pdata_{\textrm{OOD}} (\vx)$ at test-time, where the latter can have an arbitrary number of classes and is not accessible during training. Then, an OOD detector typically implements a $(C+1)$~classifier using the task's $p_{\vlambda}(y | \vx)$ with or without outlier exposure to separate $\pdata (\vx)$ and $\pdata_{\textrm{OOD}} (\vx)$ using an additional OOD class.

However, this conventional formulation does not account for assumptions in~(\ref{eq:p2}). If the empirical $\train$ with $\ptrain (\vx)$ density does not approximate true test-time $\pdata (\vx)$, the learned predictions $p_{\vlambda}(y | \vx)$ cannot be reliable due to a distributional shift. Then, test-time misclassifications are caused by a mismatch between $\ptrain (\vx)$ and a-priori inaccessible $\pdata (\vx)$. Similarly, $\pdata_{\textrm{OOD}} (\vx)$ is a result of unavailable at train-time open-world data distribution. This is sketched in Figure~\ref{fig:problem} bottom right corner: the $\pdata (\vx)$ tail is misclassified and, concurrently, there are novel OOD classes from $\pdata_{\textrm{OOD}} (\vx)$. Lastly, the statistical objective~(\ref{eq:p2}) typically cannot be fully achieved even for available $\train$ due to model underfitting $\left(\mathcal{L}_{\textrm{train}} (\vlambda) > 0 \right)$.

This analysis motivates us to \textit{narrow down a definition of in-distribution data} in the realistic data setup to a distribution of correctly classified examples only. Then, the detector's objective is to assign high confidence scores only for a distribution of positives in the proximity of $\ptrain (\vx)$. In opposite, the detector has to assign low confidence scores both for the OOD density $\pdata_{\textrm{OOD}} (\vx)$ and the misclassified data distribution ($\hat{y} \ne y$). While considering a single type of negatives is widely used in prior literature, our problem statement advocates to incorporate both types of negatives during training and revisit the conventional evaluation setup.

\subsection{Normalizing Flow Framework}\label{subsec:nflow}
Unlike other generative models, normalizing flows introduced by~\citet{rezende15} can estimate the \textit{exact data likelihoods}, which makes them an ideal candidate for IDM/OOD detection. These models use a change-of-variable formula to transform an arbitrary probability density function $p(\vz)$ into a base distribution with $p(\vu)$ density using a bijective invertible mapping $g: \mathbb{R}^D \rightarrow \mathbb{R}^D $. Usually, the mapping $g$ is a sequence of basic composable transformations. Then, the $\log$-likelihood of a $D$-dimensional input vector $\vz \sim p(\vz)$ can be estimated as
\begin{equation} \label{eq:f1}
	\log p_{\vtheta} (\vz) = \log p(\vu) + \sum_{l=1}^L \log \left| \det \mJ_l \right|,
\end{equation}
where a base random variable vector $\vu \in \mathbb{R}^{D}$ is from the standard Gaussian distribution $\vu \sim \mathcal{N}(\vzero, \mI)$ and the Jacobian matrices $\mJ^{D \times D}_l = \nabla_{\vz^{l-1}} g_{\vtheta_l}(\vz^l)$ can be sequentially calculated for the $l^{\textrm{th}}$ block of a model $g(\vtheta)$ with $L$ blocks.

\subsection{The Proposed FlowEneDet Model}\label{subsec:FEDmodel}
The conventional flow framework in Section~\ref{subsec:nflow} can estimate only the marginal likelihood as $\prod^D_{d=1} p_{\vtheta} (\vz_d)$. In result, the previous flow-based density estimator in~\citep{fishyscapes} is limited to likelihood estimates from positives only and bears a significant computational complexity by processing high-dimensional latent-space embedding vectors. First, we address the latter limitation by processing a low-dimensional free energy vectors $\vz \in \mathbb{R}^{D=2}$ that are related to the $\ptrain (\vx)$ of interest in Section~\ref{subsec:ene}. Second, we use an autoregressive interpretation of flows from Section~\ref{subsec:maf} and resolve the former limitation by introducing a distributional model for data positives and negatives in Section~\ref{subsec:nflowgmm}.

\subsubsection{Energy-Based Approach for Flows}\label{subsec:ene}
\citet{Grathwohl2020Your} and \citet{NEURIPS2020_f5496252} show that the scalar \textit{free energy score} $E_{\vlambda}(\vx)$ can be derived from a pretrained classifier $f_{\vlambda}(\vx)$ and it is theoretically aligned with the density of input $\ptrain(\vx)$ as
\begin{equation} \label{eq:ene0}
\ptrain (\vx) \approx p_{\vlambda}(\vx) = e^{-E_{\vlambda}(\vx)} / Z(\vlambda),
\end{equation}
where the free energy $E_{\vlambda}(\vx) = - \log \sum^C_{y=1} e^{ f_{\vlambda}(\vx)[y] }$ and $Z(\vlambda)$ is the normalizing constant (partition function).

The energy-based framework~\citep{energy} in~(\ref{eq:ene0}) is a key to relate the in-domain $\ptrain (\vx)$ from Section~\ref{subsec:motivation} with the trained classifier's density $p_{\vlambda}(\vx)$. We use this result in our flow-based detector by assigning its (\ref{eq:f1}) input vectors $\vz$ to the scalar energy of positives $E_{\vlambda}(\vx)$ and the scalar energy of IDM/OOD negatives $\bar{E}_{\vlambda}(\vx)$ as
\begin{equation} \label{eq:ene1}
	\vz = [-E_{\vlambda}(\vx); \bar{E}_{\vlambda}(\vx)]= [-E_{\vlambda}(\vx); \log(1 - e^{-{E}_{\vlambda}(\vx)})].
\end{equation}

\subsubsection{Autoregressive Interpretation of Flows}\label{subsec:maf}
The real-valued non-volume preserving (RNVP) architecture~\citep{45819} is a sequence coupling blocks. Each $l^{\textrm{th}}$ block represents an invertible transformation $g: \mathbb{R}^D \rightarrow \mathbb{R}^{D-d}$ for the first $d<D$ elements of vector $\vz$ as
\begin{equation} \label{eq:fed1}
	\vz^l_{1:d} = \vz^{l-1}_{1:d},~\vz^l_{d:D} = \vz^{l-1}_{d:D} \odot e^{s \left( \vz^{l-1}_{1:d} \right)} + t \left(\vz^{l-1}_{1:d} \right),
\end{equation}
where $s(\cdot)$ and $t(\cdot)$ are scale and translation operations that are implemented as two feedforward neural networks with $\vtheta$ parameters, and $\odot$ is the Hadamard (element-wise) product.

The Jacobian of such transformation is a triangular matrix with a tractable $\log$-determinant in~(\ref{eq:f1}). Importantly, \citet{maf} show that the RNVP coupling implements \textit{a special case of autoregressive transformation}. The autoregressive characterization of the coupling block using a single Gaussian is given by $m^{\textrm{th}}$ conditional likelihoods
\begin{equation} \label{eq:fed2}
	p_{\vtheta}(z^l_m|\vz^l_{1:m-1}) = \mathcal{N}(z^l_m | t_m, e^{2s_m}),
\end{equation}
where $t_m = s_m = 0$ for $\forall m \leq d$ and depend on $\vz^{l-1}_{1:d}$ only.

\begin{figure*}[t]
	\centering
	\includegraphics[width=0.95\textwidth]{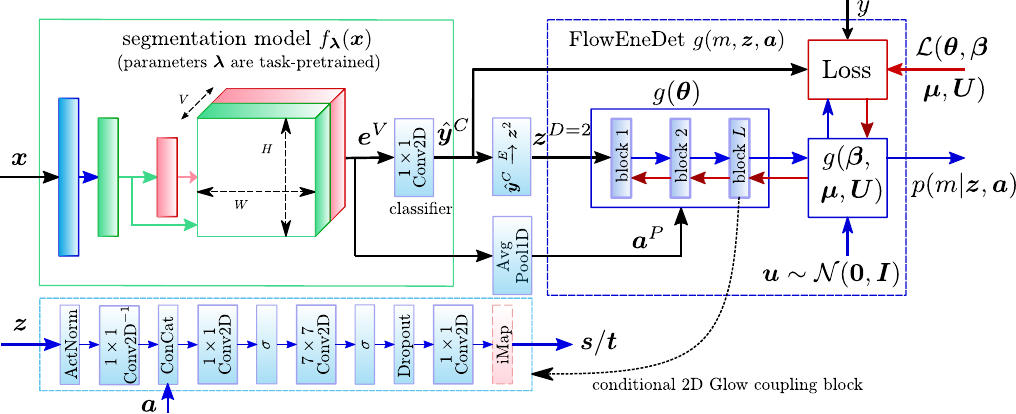}
	\caption{A pretrained segmentation model $f_{\vlambda}(\vx)$ (top left) is a multi-scale network with a linear classifier and fixed parameters $\vlambda$. Its outputs are latent-space vectors $\ve$ and unnormalized logits $\hat{\vy}$. Our FlowEneDet (right) derives an energy-based input vector $\vz$ from $\hat{\vy}$ and a condition vector $\va$ from $\ve$, and processes them by a 2D Glow-like~\citep{NEURIPS2018_d139db6a} architecture with $L$ blocks and a distributional part. Then, FlowEneDet estimates conditional likelihoods $p(m|\vz,\va)$, where the $m^{\textrm{th}}$ category defines a likelihood of image $\vx$ being either a positive or negative (IDM/OOD).}
	\label{fig:FED}
\end{figure*}

Hence, our FlowEneDet model with the $\vz \in \mathbb{R}^{D=2}$ input~(\ref{eq:ene1}) can estimate conditional likelihoods of positive's and negative's free energy scores from the output of $f_{\vlambda}(\vx)$ using the autoregressive interpretation of RNVP at every coupling~(\ref{eq:fed1}). This results in a very \textit{low-complexity architecture}, because the coupling compute is $\mathcal{O}(D^2)$. In contrast, the complexity of~\citet{fishyscapes} with latent-space vectors $\ve \in \mathbb{R}^V$ is significantly higher since $V \gg 2$ as in Figure~\ref{fig:FED}. 

\subsubsection{Distributional Model with Full Covariance}\label{subsec:nflowgmm}
The conventional choice for a base distribution in~(\ref{eq:f1}) is not suitable for modeling joint probability density~(\ref{eq:fed2}) of positive and negative energy scores~(\ref{eq:ene1}). Therefore, we replace the base univariate Gaussian in (\ref{eq:f1}) by a
\begin{equation} \label{eq:gmm1}
	p(\vu) = \vbeta \odot \mathcal{N}(\vu | \vmu, \mSigma),
\end{equation}
where $\vbeta \in \mathbb{R}^D$ is a vector of probabilities to model data imbalances between positives and negatives. A mean vector $\vmu \in \mathbb{R}^D$ and a covariance matrix $\mSigma \in \mathbb{R}^{D \times D}$ parameterize multivariate Gaussian distribution.

Then, the conditional $\log$-likelihoods $\log p_{\vtheta}(\vz|m)$ define whether an input is from positive or negative category. They can be derived by substituting (\ref{eq:gmm1}) to (\ref{eq:f1}) and conditioning each term in~(\ref{eq:fed2}) by the category $m$ using the chain rule for autoregressive output~\citep{maf} as
\begin{equation} \label{eq:gmm2}
	\begin{split}
		\log p_{\vtheta}(\vz|m) = \sum\nolimits^{D}_{d=1}  \log p_{\vtheta} (z_d | \vz_{1:d-1}, m) = \\
		\log \vbeta + \log \mathcal{N}(\vu | \vmu, \mSigma) + \sum\nolimits_{l=1}^L \log \left| \det \mJ_l \right|,
	\end{split}
\end{equation}
where the compute-intensive distributional and Jacobian terms are calculated only once for the whole model.

We model full covariance matrix $\mSigma$ of the multivariate Gaussian distribution by an upper triangular matrix $\mU$ using the Cholesky decomposition similarly to~\citep{kruse}. Then, the distributional term in~(\ref{eq:gmm2}) is given by
\begin{equation} \label{eq:gmm2-2}
	\log \mathcal{N}(\vu | \vmu, \mSigma) = \sum\nolimits_{d=1}^D \textrm{diag} (\mU)_d - \frac{1}{2} \lVert \mU (\vz-\vmu) \rVert^2_2.
\end{equation}

Using the Bayes rule for~(\ref{eq:gmm2}), confidence scores of interest can be estimated as conditional likelihoods
\begin{equation} \label{eq:gmm3}
	p_{\vtheta} (m | \vz) = p_{\vtheta} (\vz | m) p(m)  / \sum\nolimits_{d=1}^D p_{\vtheta} (\vz | m = d).
\end{equation}

Unlike the discriminative model (\ref{eq:p2}) that learns only conditionals $p_{\vlambda} (y | \vx)$, our generative FlowEneDet models the joint density $p_{\vtheta} (m, \vz)$. Hence, it exactly estimates $p_{\vtheta} (m | \vz)$ and approximates $\ptrain (\vx)$~\citep{nalisnick19b}. The joint modeling can be practically used to generate hard cases by the virtual outlier synthesis~\citep{du2022vos}.

\section{FlowEneDet for Semantic Segmentation}\label{sec:FED}
In this section, we present FlowEneDet architecture adopted for semantic segmentation. It contains two high-level parts: a sequence of $L$ coupling blocks $g(\vtheta)$ and a distributional part $g(\vbeta, \vmu, \mU)$ as shown Figure~\ref{fig:FED} (right). Next, we explain key modifications to the theoretical model from Section~\ref{subsec:FEDmodel}.

\subsection{The Proposed Architecture}\label{subsec:arch}
First, we extend the conventional RNVP coupling by 2-dimensional processing as shown in Figure~\ref{fig:FED} (bottom left). This captures information encoded along spatial dimensions for image segmentation. It is achieved by a sequence of Conv2D layers with kernels of size $1\times1$ $\rightarrow$ $\sigma$ $\rightarrow$  $7\times7$ $\rightarrow$ $\sigma$ $\rightarrow$ $1\times1$, where $\sigma$ is the sigmoid activation function.

Second, we extend the RNVP coupling by the activation normalization (ActNorm) and invertible $1 \times 1$ convolution (Conv2D$^{-1}$), which, effectively, results in a 2D Glow~\citep{NEURIPS2018_d139db6a} coupling block. Empirical experiments show that such layers significantly speed up convergence time and training stability. Dropout with 20\% probability is applied before the last $1 \times 1$ Conv2D layer to decrease overfitting. Optionally, we add an invertible map-based attention layer (iMap) from~\citep{Sukthanker_2022_CVPR}. In particular, we apply it only to the SegFormer~\citep{xie2021segformer} backbone. We empirically find that this improves training stability and decreases variance in results.

Third, we recognize that the logit-space energy score alone can limit the expressiveness of our density estimator. Therefore, we augment (condition) each coupling block by the low-dimensional embedding vector $\va^P$. A mapping from the embedding $\ve^V$ ($\ve^V \rightarrow \va^P$) is accomplished using 1D average pooling. Then, we follow~\citep{ardizzone2019guided} and concatenate $\vz$ intermediate results with the pooled projection $\va$ in Figure~\ref{fig:FED} (bottom left). We compare FlowEneDet (FED) that is configured with conditional vector $\va$ (FED-C) as well as unconditional FED-U model in our experiments.

Fourth, we improve experimental results by reparameterizing the scale operation $s(\vz_{1:d})$ in~(\ref{eq:fed1}) and the corresponding Jacobian. Particularly, we define the scale as $1 - \textrm{sigmoid}(\vz_{1:d})$ and $\log \left| \det \mJ \right| = - \textrm{softplus}(\vz_{1:d})$, which limits their range to $(0:1)$ and $(- \infty :0)$, respectively. At the same time, we follow the conventional channel-wise input masking~\citep{45819} and exchange the first and second halves of the input $\vz$ after every coupling block.

\subsection{Optimization Objective}\label{subsec:objective}
We are interested in modeling and estimating likelihoods of energy scores for positive and negative (IDM/OOD) examples as defined in~(\ref{eq:ene1}). Energy score calculation is implemented using numerically-stable \textit{logsumexp} operation. Therefore, the proposed FlowEneDet explicitly estimates conditional likelihoods $p_{\vtheta} (m | \vz, \va)$, where $m \in \{ 1, 2 \}$, $\vz \in \mathbb{R}^2$ and $\va \in \mathbb{R}^P$. At test-time, we always output likelihood of the second negative category ($m=2$) as an uncertainty estimate.

In total, FlowEneDet contains $\vtheta_{\textrm{FED}}= [\vtheta, \vbeta, \vmu, \mU]$ parameters, where $\vtheta$ are the coupling parameters and the rest of parameters describe the distributional model. All parameters are jointly optimized using an objective that maximizes~(\ref{eq:gmm3}) $\log$-likelihoods with conditional and marginal (denominator) terms. This can be simplified by a numerically-stable $\log$-softmax operation as
\begin{equation} \label{eq:sfd1}
	\mathcal{L}(\vtheta_{\textrm{FED}}) = -\sum_{i\in\sN} \log \mathrm{softmax}~\log p_{\vtheta_{\textrm{FED}}}(\vz_i, \va_i | m_i) / N,
\end{equation}
where this objective is equivalent to the cross-entropy loss.

FlowEneDet labels $m$ are binary (positive or negative examples) in the~(\ref{eq:sfd1}) loss function. We derive binary labels from the task ground-truth $y$ (including the void class) and task classifier predictions $\hat{y}$ such that $m_i = (y_i \neq \hat{y}_i$). In order to increase training stability, we optimize distributional parameters $\vbeta$ and $\textrm{diag} (\mU)$ using the same sigmoid/softplus reparameterization as for $s(\cdot)$ operation in Section~\ref{subsec:arch}.

\section{Experiments}
\label{sec:eval}
\subsection{Experimental Setup}
\label{subsec:setup}
\textbf{Task models.} We experiment with SegFormer-B2 (SF-B2)~\citep{xie2021segformer} and DeepLabV3+~\citep{deeplabv3plus2018} with ResNet-101 backbone (DL-R101) semantic segmentation models. We use their public checkpoints pretrained on Cityscapes~\citep{cityscapes}, and our code extends open-source MMSegmentation~\citep{mmseg2020} library.

\textbf{Benchmarks.} Cityscapes (CS) contains 19 labeled classes and the unlabeled void class (background). The pretrained DL-R101 and SF-B2 models achieve, correspondingly, 81.0\% and 81.1\% mean intersection over union (mIoU) metric for CS validation split. In our IDM/OOD experiments, we use 19 in-domain (ID) classes for IDM detection and the void class for OOD detection. We evaluate detection robustness by adding test-time image corruptions to CS. We follow recent robustness benchmarks~\citep{croce2021robustbench, michaelis2020benchmarking} and experiment with a synthetically-corrupted CS-C validation dataset. In particular, we apply motion blur, brightness and snow types of image corruptions with severity range from 1 (low) to 4 (high) and average corresponding results. Lastly, we use Fishyscapes~\citep{fishyscapes} (FS) and SegmentMeIfYouCan~\citep{segmentmeifyoucan} (SMIYC) benchmarks designed for OOD-only evaluations with the binary ID/OOD labels.

\textbf{Baselines.} We reimplement MSP, MLG, SML, ENE, and MCD baselines from Table~\ref{tab:landscape}. In our MCD implementation, we apply dropout only to the last linear layer to avoid high complexity. We exclude ODIN because it requires test-time gradients and underperforms compared to ENE method. We report results for other relevant baselines from Table~\ref{tab:landscape} using their best publicly available benchmark results. Due to differences in architectures and implementations, we present several FlowEneDet configurations that are comparable in terms of complexity to the above methods.

\textbf{FlowEneDet.} We experiment with the following configurations: unconditional FED-U and conditional FED-C with $P=32$ latent vectors. We train each configuration four times and report evaluation's mean ($\mu$) and standard deviation ($\pm \sigma$) for every benchmark with the exception of private test splits. Reimplemented baselines have been evaluated once. Each detector with DL-R101 and SF-B2 backbone has $L=4$ and $L=8$ coupling blocks, respectively. In addition, we can apply a test-time augmentation (TTA) with $1/4\times$, $1/2\times$ and $1\times$ image resizing to SF-B2 backbone for confidence score averaging. Our TTA configuration is chosen to match inference speed of a popular WideResNet-38 (WRN-38) backbone in FS and SMIYC leaderboards.

\begin{table}[t]
	\caption{FED SF-B2 ablation study on FS L\&F \textbf{validation split}, \%. The \textbf{best} result is highlighted. Design space is defined as follows: covariance matrix $\mU$ is full or diagonal, kernel size $K$ for the flow's Conv2D layer is $3\times3$ or $7\times7$. Our default configuration: full-covariance $\mU$, $K=7\times7$, $L=8$, and $P=32$ for FED-C or $P=0$ for FED-U.}
	\label{tab:fishyscapes-ablation}
	\centering
	\small
	\begin{tabular}{c|c|c|c|cc}
		\toprule
		\multirow{2}{*}{\shortstack{Method}} & \multirow{2}{*}{\shortstack{$\mU$}} & \multirow{2}{*}{\shortstack{$K$}} & \multirow{2}{*}{\shortstack{$P$}} & \multicolumn{2}{|c}{FS L\&F}\\
		&  &  &  & AP$\uparrow$ & FPR$_{95}\downarrow$ \\
		\midrule
		FED-U       & full & 7$\times$7  &  -  &          39.90 &           18.66\\
		FED-C       & full & 7$\times$7  &  32 &          41.15 &           11.1 \\
		FED-U (TTA) & full & 7$\times$7  &  -  &          41.75 &           10.05\\
		FED-C (TTA) & full & 7$\times$7  &  32 & \textbf{56.11} &    \textbf{3.87} \\
		FED-U (TTA) & full & 3$\times$3  &  -  &          42.28 &           9.94 \\
		FED-C (TTA) & full & 3$\times$3  &  32 &          51.98 &           6.88 \\
		FED-U (TTA) & diag & 7$\times$7  &  -  &          41.71 &           9.99 \\
		FED-C (TTA) & diag & 7$\times$7  &  32 &          51.62 &           4.04 \\
		\bottomrule
	\end{tabular}
\end{table}

\begin{table*}[h]
	\caption{OOD results for Fishyscapes \textbf{validation split}, \%. The \textbf{best} and the \textun{second best} results are highlighted.}
	\label{tab:fishyscapes-val-supp-results}
	\centering
	\small
	\begin{tabular}{c|c|c|ccc|ccc}
		\toprule
		\multirow{2}{*}{\shortstack{Method}} & \multirow{2}{*}{\shortstack{Task \\ backbone}} & CS & \multicolumn{3}{|c}{L\&F} & \multicolumn{3}{|c}{Static} \\
		&  & mIoU$\uparrow$ & AuROC$\uparrow$ & AP$\uparrow$ & FPR$_{95}\downarrow$ & AuROC$\uparrow$ & AP$\uparrow$ & FPR$_{95}\downarrow$\\
		\midrule 
		MCD     & DL-R101 & 80.3 & 88.94 & 10.85 & 37.79 & 93.14 & 25.59 & 27.24 \\
		MSP     & DL-R101 & 80.3 & 86.99 &  6.02 & 45.63 & 88.94 & 14.24 & 34.10 \\
		MLG     & DL-R101 & 80.3 & 92.00 & 18.77 & 38.13 & 92.80 & 27.99 & 28.50 \\
		ENE     & DL-R101 & 80.3 & 93.50 & 25.79 & 32.26 & 91.28 & 31.66 & 37.32 \\
		SML     & DL-R101 & 80.3 & 96.88 & 36.55 & 14.53 & \textun{96.69} & \textun{48.67} & \textbf{16.75} \\
		SynthCP & DL-R101 & 80.3 & 88.34 &  6.54 & 45.95 & 89.90 & 23.22 & 34.02 \\
		Synboost& DL-R101 & 80.3 & 94.89 & \textun{40.99} & 34.47 & 92.03 & 48.44 & 47.71 \\
		FED-U   & DL-R101 & 81.0 & \textun{97.65}\tiny$\pm$0.2 & 37.05\tiny$\pm$0.6 & \textun{11.35}\tiny$\pm$0.5 & 95.96\tiny$\pm$0.2 & 46.32\tiny$\pm$0.4 & 20.15\tiny$\pm$1.3 \\
		FED-C   & DL-R101 & 81.0 & 96.34\tiny$\pm$0.2 & 28.71\tiny$\pm$2.5 & 18.48\tiny$\pm$1.5 & 92.89\tiny$\pm$0.4 & 25.34\tiny$\pm$4.7 & 32.69\tiny$\pm$1.4 \\
		FED-U   & SF-B2   & 81.1 & 96.72\tiny$\pm$0.2          & 39.90\tiny$\pm$0.7          & 18.66\tiny$\pm$1.5          & \textbf{96.84}\tiny$\pm$0.1 & \textbf{55.93}\tiny$\pm$0.7 & \textun{17.15}\tiny$\pm$0.9 \\
		FED-C   & SF-B2   & 81.1 & \textbf{98.28}\tiny$\pm$0.1 & \textbf{42.15}\tiny$\pm$0.4 & \textbf{11.10}\tiny$\pm$0.1 & 93.31\tiny$\pm$0.8          & 47.56\tiny$\pm$2.5          & 37.53\tiny$\pm$3.1 \\	
		\midrule
		SML         & DL-WRN38 & 81.4 & 94.97                       & 22.74                       & 33.49                       & \textun{97.25}              & \textbf{66.72}                       & \textun{12.14} \\
		SynBoost    & DL-WRN38 & 81.4 & 96.21                       & \textbf{60.58}              & 31.02                       & 95.87                       & 66.44              & 25.59 \\
		FED-U (TTA) & SF-B2    & 81.1 & \textun{97.83}\tiny$\pm$0.1 & 41.75\tiny$\pm$0.3          & \textun{10.05}\tiny$\pm$0.2 & \textbf{98.30}\tiny$\pm$0.1 & \textun{66.60}\tiny$\pm$0.2 & \textbf{8.94}\tiny$\pm$0.1 \\
		FED-C (TTA) & SF-B2    & 81.1 & \textbf{99.11}\tiny$\pm$0.1 & \textun{56.11}\tiny$\pm$4.4 & \textbf{3.87}\tiny$\pm$0.2  & 96.88\tiny$\pm$0.2 & 52.61\tiny$\pm$1.4          & 14.91\tiny$\pm$1.2 \\
		\bottomrule
	\end{tabular}
\end{table*}

\textbf{Metrics.} We use standardized metrics for FS and SMIYC benchmarks: area under the receiver operating characteristic curve (AuROC), average precision (AP)~\citep{hendrycks17baseline}, and false positive rate when the true positive rate is 95\% (FPR$_{95}$)~\citep{liang2018enhancing}. The latter metric is considered the most important in practice. We use an open-mIoU metric for concurrent IDM/OOD detection evaluations. First, we compute a detection threshold using $F_1$-score~\citep{f1}. Then, this threshold is used to predict a binary (positive or IDM/OOD negative) decision. Next, the predicted negatives are added as an extra void class to IoU computation as proposed by ~\citet{grcic22eccv}. Finally, we calculate the open-mIoU metric for $(C+1)$ IoUs with averaging by $C$ classes to conform with the open-world setup. Unlike it, the conventional mIoU rejects all OOD (unlabeled void) pixels using the ground truth mask, which leads to an unrealistic closet-set recognition setup.

\subsection{Quantitative Results}
\label{subsec:quant_eval}
\textbf{Ablation study.} Table~\ref{tab:fishyscapes-ablation} shows an ablation study for FED variants with SF-B2 backbone on FS L\&F validation dataset. We find that TTA significantly increases performance metrics both for FED-U and FED-C. Next, we verify that the full covariance matrix $\mU \in \mathbb{R}^{2 \times 2}$ from Section~\ref{subsec:nflowgmm} outperforms the univariate $\textrm{diag} (\mU) \in \mathbb{R}^{2}$ approach. Finally, a $7 \times 7$ kernel size with larger receptive field is superior to a $3 \times 3$ Conv2D layer for a more advanced FED-C configuration. We use the selected configurations as default in further experiments. Appendix contains an extended ablation study.

\textbf{OOD-only detection.} Tables~\ref{tab:fishyscapes-val-supp-results}-\ref{tab:smiyc-supp-results} present comprehensive OOD evaluations when applied to FS public validation split as well as FS and SMIYC private test splits, respectively. Our conditional FED-C configuration exceeds or is on par with the state-of-the-art in majority of metrics for the setup without task retraining, and even outperforms the best methods with retraining (NFlowJS, DenseHybrid, PEBAL) in Table~\ref{tab:fishyscapes-test-supp-results} on FS L\&F test split (50.15\% AP, 5.2\% FPR$_{95}$).

The only outlier is FS Static dataset in Tables~\ref{tab:fishyscapes-val-supp-results}-\ref{tab:fishyscapes-test-supp-results}, where unconditional FED-U is consistently superior than the more advanced FED-C variant. Particularly, FED-U has the second best test split results using AP metric (67.80\% AP for FED-U vs. 72.59\% AP for SynBoost), but it underperforms in FPR$_{95}$ (21.58\% FPR$_{95}$ for FED-U vs. 17.43\% FPR$_{95}$ for flow embedded density method~\citep{fishyscapes}). A possible reason why FED-C achieves lower performance metrics on FS Static than the FED-U is the distribution of latent-space features in OOD objects that cannot be properly captured by our na\"ive average pooling. Hence, a more robust feature pooling can be a topic for future research.

\begin{table*}[h]
	\caption{OOD results for Fishyscapes \textbf{test split}, \%. The \textbf{best} and the \textun{second best} results are highlighted.}
	\label{tab:fishyscapes-test-supp-results}
	\centering
	\small
	\begin{tabular}{c|c|c|c|cc|cc}
		\toprule
		\multirow{2}{*}{\shortstack{Method}} & \multirow{2}{*}{\shortstack{Intact mIoU, \\ no retraining}} & \multirow{2}{*}{\shortstack{Task \\ backbone}} & CS & \multicolumn{2}{|c}{L\&F} & \multicolumn{2}{|c}{Static} \\
		&  &  & mIoU$\uparrow$ & AP$\uparrow$ & FPR$_{95}\downarrow$ & AP$\uparrow$ & FPR$_{95}\downarrow$\\
		\midrule
		MSP                            & \cmark & DL-R101  & 80.3 & 1.77           & 44.85          & 12.88 & 39.83 \\
		Emb. density                   & \cmark & DL-R101  & 80.3 & 4.25           & 47.15          & 62.14 & \textbf{17.43} \\
		SML                            & \cmark & DL-R101  & 80.3 & 31.05          & 21.52          & 53.11 & 19.64 \\
		Image Resynthesis              & \cmark & PSP-R101 & 79.9 & 5.70           & 48.05          & 29.60 & 27.13 \\
		SynBoost                       & \cmark & DL-WRN38 & 81.4 & \textun{43.22} & 15.79          & \textbf{72.59} & \textun{18.75} \\
		FED-U TTA                      & \cmark & SF-B2    & 81.1 & 20.45          & \textun{11.38} & \textun{67.80} & 21.58 \\
		FED-C TTA                      & \cmark & SF-B2    & 81.1 & \textbf{50.15} & \textbf{5.20}  & 61.06 & 31.97 \\
		\midrule
		NFlowJS                        & \xmark & LDN-121  & 77.4 & 43.66          & 8.61           & 54.68 & 10.00 \\
		DenseHybrid                    & \xmark & DL-WRN38 & 81.0 & 43.90          & 6.18           & 72.27 & 5.51 \\
		PEBAL                          & \xmark & DL-WRN38 & 80.7 & 44.17          & 7.58           & 92.38 & 1.73 \\
		GMMSeg                         & \xmark & DL-R101  & 81.1 & 55.63          & 6.61           & 76.02 & 15.96 \\
		\bottomrule
	\end{tabular}
\end{table*}

\begin{table*}[ht]
	\caption{OOD results for SMIYC \textbf{test split}, \%. The \textbf{best} and the \textun{second best} results are highlighted.}
	\label{tab:smiyc-supp-results}
	\centering
	\small
	\begin{tabular}{c|c|c|c|cc|cc}
		\toprule
		\multirow{2}{*}{\shortstack{Method}} & \multirow{2}{*}{\shortstack{Intact mIoU, \\ no retraining}} & \multirow{2}{*}{\shortstack{Task \\ backbone}} & CS & \multicolumn{2}{|c}{Obstacle} & \multicolumn{2}{|c}{L\&F} \\
		&  &  & mIoU$\uparrow$ & AP$\uparrow$ & FPR$_{95}\downarrow$ & AP$\uparrow$ & FPR$_{95}\downarrow$\\
		\midrule
		MSP              & \cmark & DL-WRN38 & 81.4 & 15.7 & 16.6 & 30.1 & 33.2 \\
		MCD              & \cmark & DL-R101  & 80.3 & 4.9  & 50.3 & 36.8 & 35.6 \\
		Emb. density     & \cmark & DL-R101  & 80.3 & 0.8  & 46.4 & 61.7 & 10.4 \\
		Void Classifier  & \cmark & DL-R101  & 80.3 & 10.4 & 41.5 &   4.8  & 47.0 \\
		Image Resynthesis& \cmark & PSP-R101 & 79.9 & 37.7 &  4.7 & 57.1 &  8.8 \\
		Mah. distance    & \cmark & DL-WRN38 & 81.4 & 20.9 & 13.1 & 55.0 & 12.9 \\
		SynBoost         & \cmark & DL-WRN38 & 81.4 & \textun{71.3} &  \textun{3.2}   & \textbf{81.7}  &  \textun{4.6} \\
		FED-C (TTA)      & \cmark & SF-B2    & 81.1 & \textbf{73.7} & \textbf{1.0} & \textun{79.8} & \textbf{2.9} \\
		\midrule
		NFlowJS          & \xmark & LDN-121 & 77.4  & 85.6 & 0.4 & 89.3 & 0.7 \\
		DenseHybrid      & \xmark & LDN-121 & N/A   & 81.7 & 0.2 & 78.7 & 2.1 \\
		\bottomrule
	\end{tabular}
\end{table*}

OOD detection results cannot be considered separately from the task's semantic segmentation accuracy metric itself. For example, DenseHybrid and PEBAL sacrifice, correspondingly, 0.4\% and 0.7\% Cityscapes closed-set mIoU accuracy due to a setup with retraining. A benchmark-agnostic task model with the corresponding detector that can be universally applied in all experiments using the same parameters is another important factor when analyzing Table~\ref{tab:fishyscapes-test-supp-results}-\ref{tab:smiyc-supp-results} results. For instance, DenseHybrid uses convolutional DL-WRN38 backbone for FS, but LDN-121 backbone for SMIYC. On the other hand, PEBAL trains several detector models with different hyperparameters and applies each checkpoint depending on the selected benchmark. In our empirical studies, we find that the transformer-based SegFormer-B2 is more universally-applicable segmentation backbone. As a result, we apply the same backbone and detector parameters using a single checkpoint file to all our OOD evaluations without additional hyperparameter tuning.

\begin{table}[t]
	\caption{Concurrent IDM/OOD detection for Cityscapes (CS), corrupted CS-C and snow-only CS-C, open-mIoU \%.}
	\label{tab:cityscapes-results}
	\centering
	\small
	\begin{tabular}{c|c|c|c|c}
		\toprule
		Method & Backbone & CS$\uparrow$ & CS-C$\uparrow$ & Snow$\uparrow$ \\
		\midrule
		None & DL-R101 & 81.0 & 53.8 & 15.4 \\
		MCD  & DL-R101 & 54.6 & 39.3 & 15.3 \\
		MSP  & DL-R101 & 61.7 & 44.3 & 16.5 \\
		ENE  & DL-R101 & 52.8 & 38.7 & 17.1 \\
		SML  & DL-R101 & 84.4 & 57.4 & 12.6 \\
		FED-U& DL-R101 & \textbf{84.6}\tiny$\pm$0.6 & \textbf{59.9}\tiny$\pm$0.8 & \textbf{18.9}\tiny$\pm$1.5 \\
		\midrule
		None& SF-B2 & 81.1 & 62.21 & 35.83 \\
		MCD & SF-B2 & 58.1 & 47.77 & 30.33 \\
		MSP & SF-B2 & 63.6 & 52.22 & 32.82 \\
		ENE & SF-B2 & 68.9 & 59.91 & 45.63 \\
		SML & SF-B2 & 81.4 & 66.26 & 40.30 \\
		FED-U & SF-B2 & 81.4\tiny$\pm$0.8 & 70.1\tiny$\pm$0.7 & 51.5\tiny$\pm$0.8 \\
		FED-C (TTA) & SF-B2 & \textbf{86.6}\tiny$\pm$0.6 & \textbf{74.5}\tiny$\pm$0.8 & \textbf{54.3}\tiny$\pm$0.6 \\
		\bottomrule
	\end{tabular}
\end{table}

\textbf{Concurrent IDM/OOD detection.} Table~\ref{tab:cityscapes-results} demonstrates the utility of concurrent IDM/OOD detection on CS, CS-C and snow-only corruption using the open-mIoU metric. Our FED-U marginally outperforms SML on the uncorrupted CS because OOD pixels represent the majority of negatives in this case, while FED-C with TTA averaging and $P=128$ achieves 5.2\% higher result. However, an amount of IDM negatives increases in case of CS-C and, especially, snow-type corruption. Then, OOD-focused SML is inferior even when compared to ENE~\citep{NEURIPS2020_f5496252}. Our FED-U surpasses others by a larger margin (2.5-6\% open-mIoU) on CS-C and snow-only case relatively to the uncorrupted CS (up to 0.2\% improvement), while a more complex FED-C with TTA shows an additional 3-4\% gain. 

Interestingly, transformer-based SF-B2 is significantly more robust (10-30\% higher open-mIoU) to corruptions than the convolutional DL-R101. Lastly, Table~\ref{tab:cityscapes-results} shows that image distortions present a significant threat to task's accuracy and not all IDM/OOD detectors are accurate enough to surpass a simple no-detector baseline using the open-mIoU metric. Therefore, it is important to use robust task's backbone \eg~transformer-based SegFormer~\citep{zhou22m} and avoid operating in an extreme environment when detector predicts broadly low-confident segmentation predictions.

\subsection{Qualitative Results}
\label{subsec:qual_eval}
\begin{figure*}[ht]
	\centering
	\includegraphics[width=1.0\textwidth]{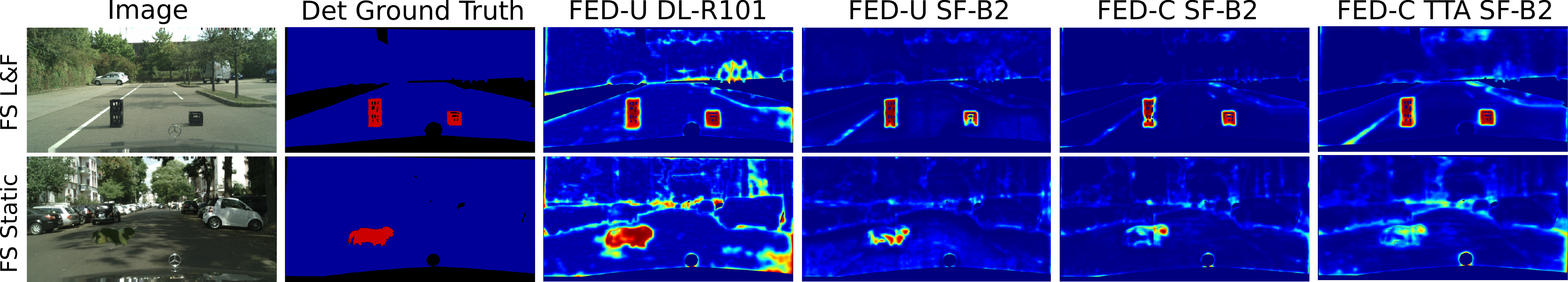}
	\caption{This figure presents: images from FS L\&F and Static validation datasets, OOD ground truth, and predictions for FED variants. FED-U predictions with DL-R101 backbone are less precise than the ones with SF-B2 transformer. Test-time augmentation (TTA) allows to further refine the exact shape of OOD objects by averaging multi-scale confidence scores.}
	\label{fig:qual}
\end{figure*}

Figure~\ref{fig:qual} compares qualitative results when different FED configurations are applied to the FS validation data. FED-U detector with the convolutional DL-R101 backbone outputs significantly less accurate confidence scores when compared to the transformer-based SF-B2 backbone. In particular, convolutional backbone produces noisy predictions for certain in-domain areas such as road patterns or a clutter of small objects in the background. We believe, this is related to a very local receptive field for convolutional backbones. FED configurations with the SF-B2 backbone output more consistent confidence scores due to global transformer receptive field. The FED configuration with TTA improves predictions by capturing very fine details in OOD object shapes. This is related to the convolutional architecture of our flow network itself, and TTA's multi-scale detection allows to partially overcome this limitation. Also, though not visible in these examples, TTA likely suppresses spurious false predictions because it smooths the estimated scores. Appendix contains additional qualitative visualizations.

\subsection{Complexity Evaluations}
\label{sec:comp_eval}
Table~\ref{tab:complexity-results} reports complexity estimates for the evaluated DL-R101 and SF-B2 task models with detectors using frames per second (fps) metric with a size-1 mini-batch on A6000 GPU and the size of all floating-point parameters. Also, we include SynBoost with WRN-38. The first row shows complexity metrics for the task model and computation-free detectors~(MSP, ENE, SML). The reimplemented MCD with 32 forward passes has a dropout layer applied only to the classifier layer to avoid high complexity.

FED detector variants contain 4 and 8 coupling blocks for DL-R101 and SF-B2, respectively. Then, its model size is marginally larger (up to 8\% for FED-C) than the task model itself. In comparison, reconstruction-based SynBoost is more than 20$\times$ larger than our FED-C with SF-B2. Inference speed without TTA is 5\% to 44\% lower depending on the backbone and architecture. The enabled TTA nearly linearly decreases inference speed in the current off-the-shelf implementation. This can be improved if exclude task's processing from TTA and apply it only to the FED detector.

\begin{table}[ht]
	\caption{Inference speed: frames per second (fps) on A6000 GPU and total model size (MB) for 1024$\times$2048 images.}
	\label{tab:complexity-results}
	\centering
	\small
	\begin{tabular}{c|c|c|c}
		\toprule
		Method & Backbone & Speed, fps$\uparrow$ & Size, MB$\downarrow$ \\
		\midrule
		MSP, ENE, SML & DL-R101 & 4.46 & 230.44 \\
		MCD & DL-R101 & 3.79 & 230.44 \\
		FED-U   & DL-R101 & 4.39 & 230.53 \\
		FED-C   & DL-R101 & 4.25 & 236.01 \\
		\midrule
		SynBoost & DL-WRN38 & 0.9 & 2,286.80 \\
		MSP, ENE, SML & SF-B2 & 5.2 & 94.47 \\
		FED-U (TTA)  & SF-B2 & 4.1 (2.2) & 94.62 \\
		FED-C (TTA)  & SF-B2 & 3.6 (0.9) & 101.69 \\
		\bottomrule
	\end{tabular}
\end{table}

\section{Conclusions}
\label{sec:conclusion}
In this paper, we analyzed a practical data setup with distributional shifts and out-of-distribution classes, which can result in critically-incorrect predictions produced by ML-based semantic segmentation models. To improve task model robustness, we proposed to incorporate a concurrent IDM/OOD detector to predict in-distribution misclassified data points and out-of-distribution classes. While IDM/OOD detection is challenging for certain types of corruptions, we significantly improved detection results using the proposed normalizing flow-based FlowEneDet model.

FlowEneDet with 2D architecture explicitly modeled likelihoods for semantic segmentation's positive (correctly classified) and negative (IDM/OOD) pixels using low-complexity energy-based inputs. We achieved promising results in IDM and/or OOD detection without task's retraining on Cityscapes, Cityscapes-C, Fishyscapes and SegmentMeIfYouCan benchmarks. This setup can extend already deployed segmentation models, keep their original mIoU accuracy intact, and improve practical open-mIoU metric. Moreover, we showed that FlowEneDet has relatively low complexity and memory overhead when applied to DeepLabV3+ and a more empirically robust SegFormer backbone.

\bibliography{paper}

\newpage
\ifreport
\appendix

\section{Implementation Details}
\label{subsec:app_hyperparameters}

\textbf{Initialization.} Convolutional parameters in the FED network $g(\vtheta)$ are initialized using the default scheme in PyTorch. ActNorm and iMap are reimplemented and initialized according to~\citep{NEURIPS2018_d139db6a, Sukthanker_2022_CVPR} references. Distributional parameters in $g(\vbeta, \vmu, \mU)$ are initialized with zero values. A subset of them $\left(\vbeta~\textrm{and}~\textrm{diag} (\mU) \right)$ are passed through a SoftPlus activation, which results in a strictly non-negative values.

\textbf{Training.} FED training phase takes only few GPU-hours and has the following hyperparameters: AdamW optimizer with initial 1e-3 learning rate, which is reduced by a factor of 10 every 15,000 iterations. We use in total 50,000 iterations and a mini-batch size of 4. In addition, a warm-up phase with the learning rate gradually increasing from 1e-6 to 1e-3 is applied during first 4,000 iterations. We select the highest learning rate from the \{1e-2, 1e-3, 1e-4\} range using ablation study. Practically, the number of training iterations can be substantially decreased (e.g. to 20,000 iterations) without a significant drop in IDM/OOD metrics. We use the default image crop sizes during training: 512$\times$1024 for DL-R101 and 1024$\times$1024 for SF-B2 backbone.

\textbf{Inference.} Inference is done on full-size images without cropping for DL-R101 task backbone. We use the reference implementation for SF-B2 backbone, where 1024$\times$1024 cropping with sliding is accomplished at test-time. Next, we discuss details about used test-time augmentation (TTA). TTA is a common technique to improve inference results for segmentation models and is available out-of-the-box in MMSegmentation library. In our case, we use TTA for input image resizing and averaging output scores without any other augmentations. We optionally apply TTA to FlowEneDet in order to increase IDM/OOD metrics at the expense of lower inference speed as reported in Section 5.4. During the training phase TTA doesn't require any modification: FED is trained by input/output tensors with $1/4 \times$ spatial dimensions of image size. In other words, the $1/4 \times$ rate is identical to the task's classifier resolution during training and inference without TTA. In case of the enabled TTA, inputs images are resized to have [$1/4 \times$, $1/2 \times$, $1 \times$] resolution, while FED input/output tensors are internally upsampled by a factor of $4 \times$ from the original $1/4\times$ resolution \ie~FED rates become [$1/4 \times$, $1/2 \times$, $1 \times$] as well. Effectively, segmentation backbone processes images with the original or downsampled resolution, while FED operates at the original or upsampled resolution \wrt~the training phase. This technique helps us to capture small- and large-scale OOD objects. A more compute-efficient approach is to train a set of multi-scale FED detectors with aggregation at the expense of marginally higher memory footprint.

\begin{table*}[ht]
	\caption{Ablation study of architectural choices for FED SF-B2 variants when applied to OOD detection on FS L\&F and Static \textbf{validation split} and IDM/OOD detection using CS \textbf{validation split}, \%. The \textbf{best} and the \textun{second best} results are highlighted. Design space is defined as follows: covariance matrix $\mU$ is full or diagonal, kernel size $K$ for the flow's Conv2D layer is $3\times3$, $7\times7$ or $11\times11$, number of coupling blocks $L$ is 4 or 8, the size $P$ of condition vector $\va$ is 32 or 128. Our default configuration: full-covariance $\mU$, $K=7\times7$, $L=8$, and $P=32$ for FED-C or $P=0$ for FED-U.}
	\label{tab:ablation-results}
	\centering
	\small
	\begin{tabular}{c|c|c|c|c|cc|cc|c}
		\toprule
		\multirow{2}{*}{\shortstack{Method}} & \multirow{2}{*}{\shortstack{$\mU$}} & \multirow{2}{*}{\shortstack{$K$}} & \multirow{2}{*}{\shortstack{$L$}} & \multirow{2}{*}{\shortstack{$P$}} & \multicolumn{2}{|c}{FS L\&F} & \multicolumn{2}{|c|}{FS Static} & CS \\
		&  &  &  &  & AP$\uparrow$ & FPR$_{95}\downarrow$ & AP$\uparrow$ & FPR$_{95}\downarrow$ & open-mIoU$\uparrow$ \\
		\midrule
		FED-U       & full & 7$\times$7  & 8 &  -  &          39.90 &          18.66 &          55.93 &         17.15  &         81.43 \\
		FED-C       & full & 7$\times$7  & 8 &  32 &          41.15 &          11.10 &          47.56 &         37.53  &         77.61 \\
		\textbf{\textit{FED-U (TTA)}} & full & 7$\times$7  & 8 &  -  &          41.75 &          10.05 & \textun{66.60} &  \textun{8.94} &         81.77 \\
		\textbf{\textit{FED-C (TTA)}} & full & 7$\times$7  & 8 &  32 & \textun{56.11} &  \textbf{3.87} &          52.61 &         14.91  &         79.40 \\
		FED-U (TTA) & full & 3$\times$3  & 8 &  -  &          42.28 &           9.94 &          65.98 &          9.09  &         81.13 \\
		FED-C (TTA) & full & 3$\times$3  & 8 &  32 &          51.98 &           6.88 &          53.98 &         13.69  &         79.14 \\
		FED-U (TTA) & full & 11$\times$11 & 8 &  -  &          40.36 &           9.98 & \textbf{66.80} &  \textbf{8.93} & \textun{82.66} \\
		FED-C (TTA) & full & 11$\times$11 & 8 &  32 & \textbf{56.84} &  \textun{4.19} &          51.47 &         16.93  &         76.34 \\
		FED-U (TTA) & diag & 7$\times$7  & 8 &  -  &          41.71 &           9.99 &          66.21 &          9.09  &         81.98 \\
		FED-C (TTA) & diag & 7$\times$7  & 8 &  32 &          51.62 &           4.04 &          55.66 &         13.15  &         81.13 \\
		FED-U (TTA) & full & 7$\times$7  & 4 &  -  &          41.57 &           9.92 &          66.21 &          9.15  &         82.00 \\
		FED-C (TTA) & full & 7$\times$7 & 4 &  32 &          49.54 &           4.63 &          50.65 &         15.89  &         71.86 \\
		FED-C (TTA) & full & 7$\times$7  & 8 & 128 &          26.00 &          17.22 &          32.57 &         22.24  & \textbf{86.59} \\
		\bottomrule
	\end{tabular}
\end{table*}

\section{Extended Ablation Study and Discussion on Limitations}
\label{subsec:app_ablation}

Table~\ref{tab:ablation-results} presents an ablation study of various architectural tradeoffs for FED detector with SF-B2 backbone. We choose a more robust SF-B2 here instead of DL-R101 backbone because the latter shows similar trends on average, but has significantly higher metric's variances. Specifically, we evaluate: unconditional FED-U and conditional FED-C, full or diagonal covariance matrix $\mU$, kernel size $K$ ($3\times3$, $7\times7$ or $11\times11$) for the flow's Conv2D layer that defines spatial receptive field, number of coupling blocks $L$ (4 or 8), and the length $P$ of condition vector $\va$ (32 or 128).

Note that the open-mIoU evaluation in Table~\ref{tab:ablation-results} is different for the configuration with TTA and without TTA. The configurations without TTA are implemented exactly as described in Section 5.1 with the closed-set mIoU of 81.1\%. However, IDM detection is not feasible for the multi-scale processing scheme described in Appendix~\ref{subsec:app_hyperparameters}, where the backbone and FED network are trained by inputs with a certain resolution scheme ($1\times$ and $1/4\times$, respectively), but tested with another resolution setup [$1/4\times$, $1/2\times$, $1\times$] both for backbone and FED network). Therefore, we derive a modified multi-scale scheme from the reference scheme for SegFormer TTA in MMSegmentation. During inference with the enabled TTA for open-mIoU evaluation in Table~\ref{tab:ablation-results}, the backbone input rate ([$1/2\times$, $1\times$, $3/2\times$]) is consistent with the FED input rate [$1/8\times$, $1/4\times$, $3/8\times$]. Hence, we preserve the same $1/4\times$ rate for the FED network during train and inference phases to successfully detect misclassifications. This TTA scheme increases closed-set mIoU from 81.1\% to 81.75\%. For reference, we report modified OOD scores for this TTA scheme on FS validation dataset using [AuROC, AP, FPR$_{95}$] format:
\begin{itemize}
	\small
	\itemsep0em
	\item FED-U L\&F: [97.83$\rightarrow$98.51, 41.75$\rightarrow$49.03, 10.05$\rightarrow$7.66]
	\item FED-C L\&F: [99.11$\rightarrow$99.27, 56.11$\rightarrow$52.92, 3.87$\rightarrow$2.95]
	\item FED-U Stat: [98.30$\rightarrow$97.80, 66.60$\rightarrow$66.53, 8.94$\rightarrow$10.31]
	\item FED-C Stat: [96.88$\rightarrow$95.51, 52.61$\rightarrow$52.78, 14.91$\rightarrow$25.63]
\end{itemize}

In our ablation study in Table~\ref{tab:ablation-results}, we verify that the full covariance matrix $\mU \in \mathbb{R}^{2 \times 2}$ outperforms the univariate $[\textrm{diag} (\mU)] \in \mathbb{R}^{2}$ approach in most cases. Similarly, the higher number of coupling blocks $L$ results in better metrics. A $11 \times 11$ kernel size with larger receptive field is superior than our default $7 \times 7$ Conv2D layer in most cases. So, our default choice is suboptimal in the sense of performance metrics, but better in terms of inference speed and memory footprint. A transformer architecture with the global attention for the flow network can be an interesting future direction~\citep{Sukthanker_2022_CVPR} to resolve a problem with the limited receptive field in convolutional layers.

\begin{figure*}[th]
	\centering
	\includegraphics[width=0.98\textwidth]{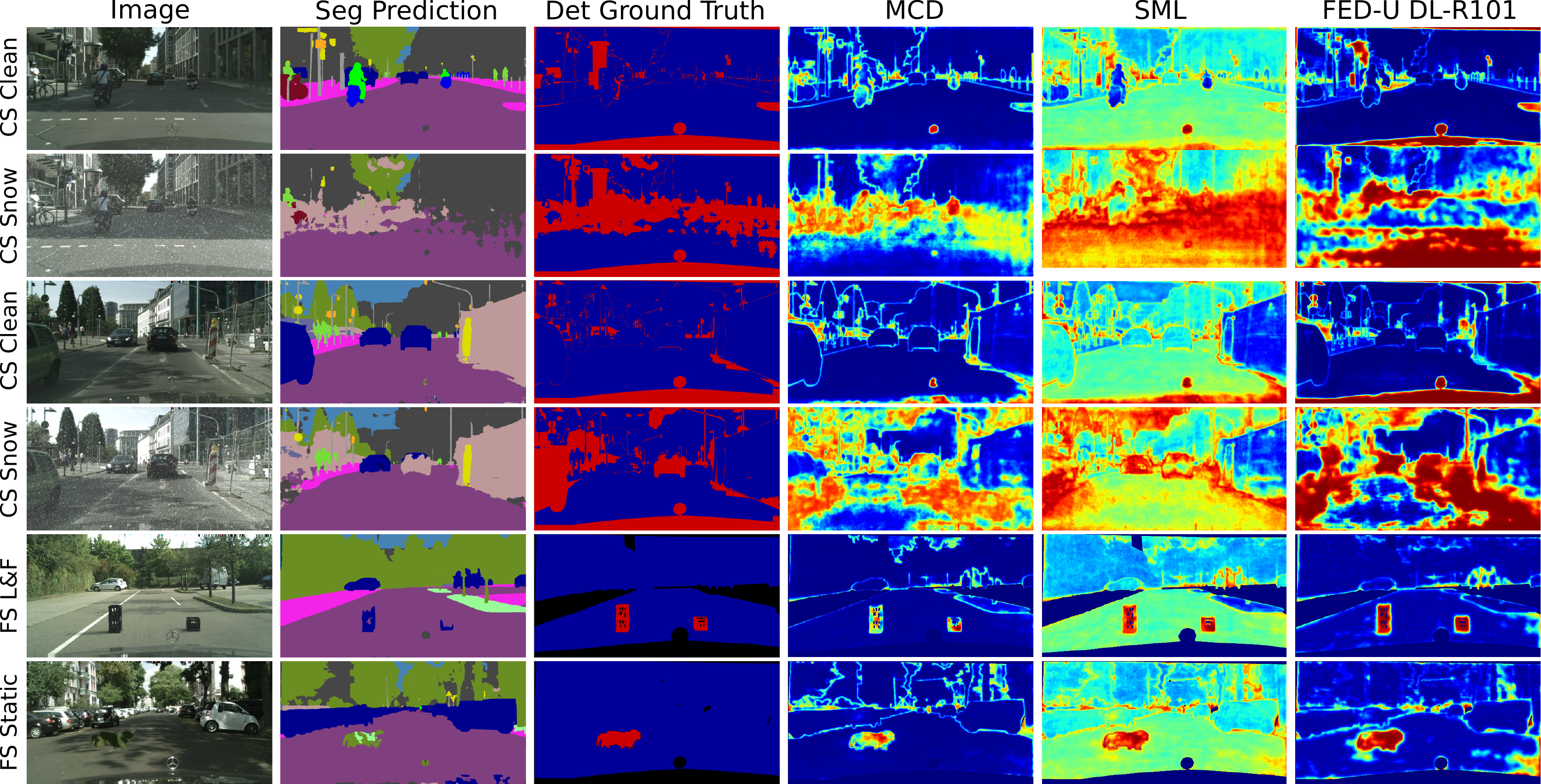}
	\caption{This figure shows from left to right: input image, DL-R101 segmentation prediction, IDM/OOD detection ground truth, and detection predictions for MCD~\citep{ken}, SML~\citep{Jung_2021_ICCV} and our FED-U detector. Each input image example is from the corresponding validation dataset, specifically, from top to bottom: two Cityscapes (CS) images and the same images corrupted by the snow corruption from Cityscapes-C, an image from the Fishyscapes (FS) L\&F and Static validation splits. Detector's task is to predict IDM/OOD pixels as red scores and correctly classified pixels as blue scores. Black area represents an ignored void class in FS datasets. Compared to other detectors, our FED-U separates IDM/OOD pixels more accurately. At the same time, IDM/OOD detection is quite challenging for heavily corrupted environment such as the snowy weather when the predicted segmentation becomes very imprecise.}
	\label{fig:qual-supp}
\end{figure*}

The length $P$ of the condition vector $\va^P$ in the current FED-C plays an ambivalent role. The larger ($P=128$) produces an excellent CS open-mIoU (86.59\%) compared to the configuration with $P=32$ (79.4\%), but significantly underperforms in FS benchmark (17.22\% FPR$_{95}$ vs. 3.87\% FPR$_{95}$ for FS L\&F). At the same time, the unconditional FED-U (\ie~$P=0$) outperforms FED-C with $P=32$ in FS Static and CS open-mIoU. Therefore, we observe that the most simplistic compute-free average pooling technique in FED-C model achieves state-of-the-art results in FS L\&F and SMIYC, but underperforms in FS Static and CS's open-mIoU due to, possibly, two different reasons. We hypothesize that a larger $P$ improves in-domain density estimation because latent-space embeddings contain more information about feature distribution, which is reflected in the excellent CS open-mIoU metric. At the same time, out-of-domain data can have a significant distributional shift. It seems to be the case in FS Static split, where FED-C underperforms compared to the embedding-unconditional FED-U model. Therefore, we conclude that FED-C approach is beneficial in general in comparison to FED-U. However, its current major limitation is in the feature pooling mechanism. We believe, FED-C results can be further improved and be more consistent across multiple datasets, if the pooled condition vector $\va$ satisfies the following: a) contains sufficient latent-space information for in-domain density estimation, and b) represents features that are robust to distributional shifts. We hope these observations will inspire follow-up research.

\section{Extra Qualitative Results}
\label{sec:app_qual_supp}
Figure~\ref{fig:qual-supp} shows additional qualitative results for our most low-complexity FED-U configuration with DL-R101 as well as MCD and SML. We plot confidence scores with a normalization to [0:1] range, where red (0) and blue (1) represent the most uncertain and certain areas, respectively. Normalization statistics are derived for each dataset before plotting detection predictions.

We select two examples from the uncorrupted CS, and the corresponding CS-C validation dataset with the lowest severity snow corruption. The second column shows segmentation model predictions, and the third column highlights its correctly classified pixels (blue), the union of IDM and OOD pixel masks (red) \ie~the detection ground truth. Last two rows show images from FS L\&F and Static validation datasets. Unlike CS, FS ground truth contains only OOD pixels (red), normal objects (blue), and the ignored during evaluation void class (black).

Our detector visually better matches detection ground truth masks. Notably, SML fails in assigning high confidence scores for in-domain positives (yellow and green instead of blue), and MCD is not consistent when assigning low confidence scores for OOD areas (green and blue instead of red). Finally, we emphasize that weather corruptions \eg~snow can pose a considerable difficulty for semantic segmentation performance as well as IDM/OOD detection. Certainly, decision-critical applications have to avoid operating in such extreme environment as soon as detector signals about broadly low-confident segmentation predictions.

\fi

\end{document}